\documentclass{article}
\usepackage{spconf,amsmath,graphicx}
\usepackage{color}
\usepackage{multirow}
\usepackage{amssymb}
\usepackage{graphicx}
\usepackage{subfig}
\usepackage{rotating}

\title{Occluded Person Re-Identification \\via Relational Adaptive Feature Correction Learning}

\name{Minjung Kim, MyeongAh Cho, Heansung Lee, Suhwan Cho, Sangyoun Lee}
\address{School of Electrical and Electronic Engineering, Yonsei University}

\begin{document}

\maketitle

\begin{abstract}
Occluded person re-identification (Re-ID) in images captured by multiple cameras is challenging because the target person is occluded by pedestrians or objects, especially in crowded scenes. In addition to the processes performed during holistic person Re-ID, occluded person Re-ID involves the removal of obstacles and the detection of partially visible body parts. Most existing methods utilize the off-the-shelf pose or parsing networks as pseudo labels, which are prone to error. To address these issues, we propose a novel Occlusion Correction Network (OCNet) that corrects features through relational-weight learning and obtains diverse and representative features without using external networks. In addition, we present a simple concept of a center feature in order to provide an intuitive solution to pedestrian occlusion scenarios. Furthermore, we suggest the idea of Separation Loss (SL) for focusing on different parts between global features and part features. We conduct extensive experiments on five challenging benchmark datasets for occluded and holistic Re-ID tasks to demonstrate that our method achieves superior performance to state-of-the-art methods especially on occluded scene.
\end{abstract}

\begin{keywords}
Occluded Person Re-Identification, Relation Network, Person Re-Identification, Deep Learning
\end{keywords}

\let\thefootnote\relax\footnote{This work was supported by the Institute of Information \& communications Technology Planning \& Evaluation(IITP) grant funded by the Korean government(MSIT) (No. 2021-0-00172, The development of human Re-identification and masked face recognition based on CCTV camera)}

\section{Introduction}
\label{sec:intro}

Person re-identification (Re-ID) involves identifying people appearing in the images captured by multiple cameras with non-overlapping domains. Most of methods utilize a representation by extracting global features from complete pedestrian images. However, these approaches have limitations in addressing the occluded Re-ID, which frequently occurs in crowded and complex scenes featuring multiple obstacles.
\begin{figure}[t]
		\centering
		\includegraphics[width=0.8\columnwidth]{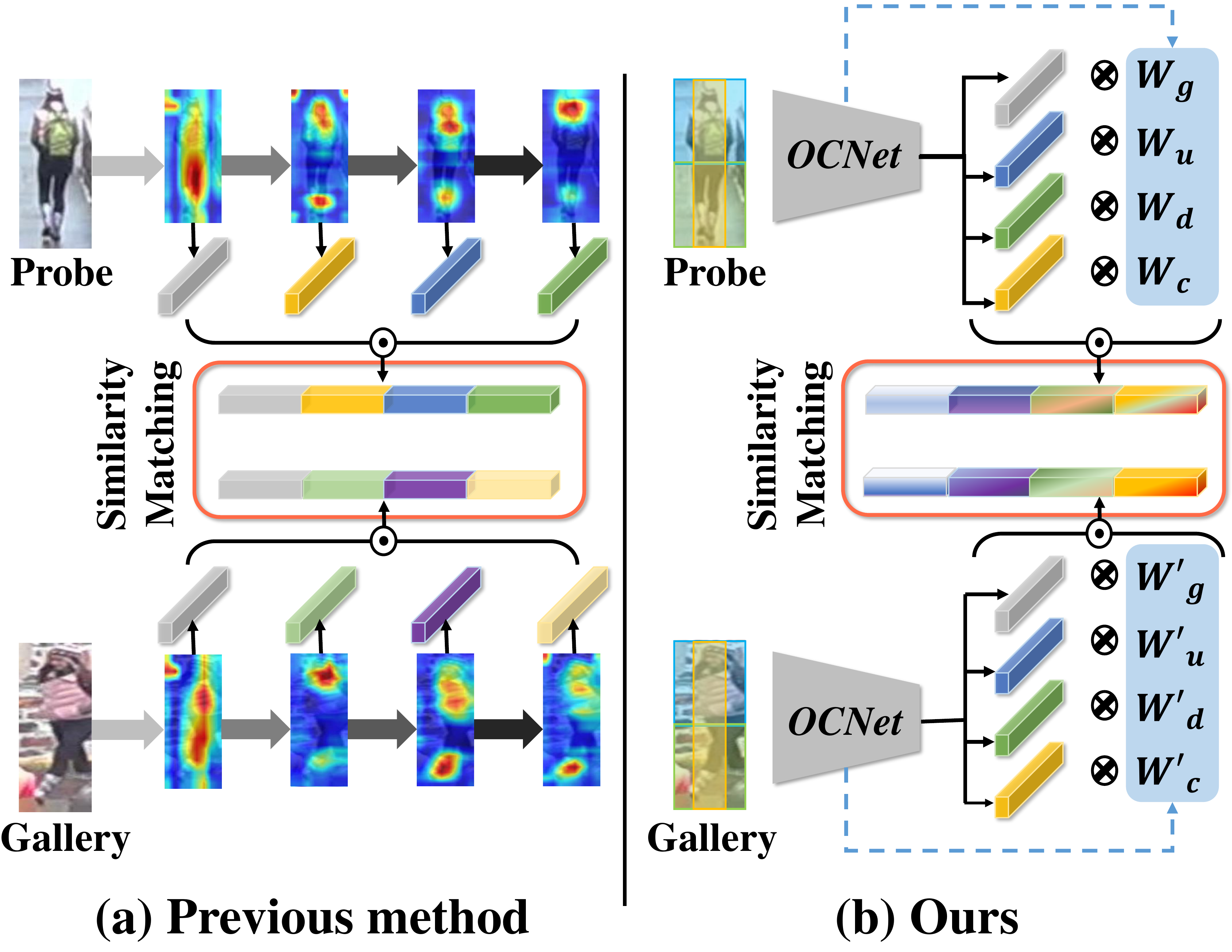}
		\caption{Challenge related to feature misalignment and our proposed OCNet. (a) Previous approaches suffered from feature misalignment problem when calculating the feature similarity because features are extracted in the order of high activation. (b) OCNet aligns various features in advance and then corrects the features with weight features $W$ obtained from relational information.}
		\label{mis}
	\end{figure}

\begin{figure*}[!t]
		\centering
		\includegraphics[width=0.8\linewidth, height=4.3cm]{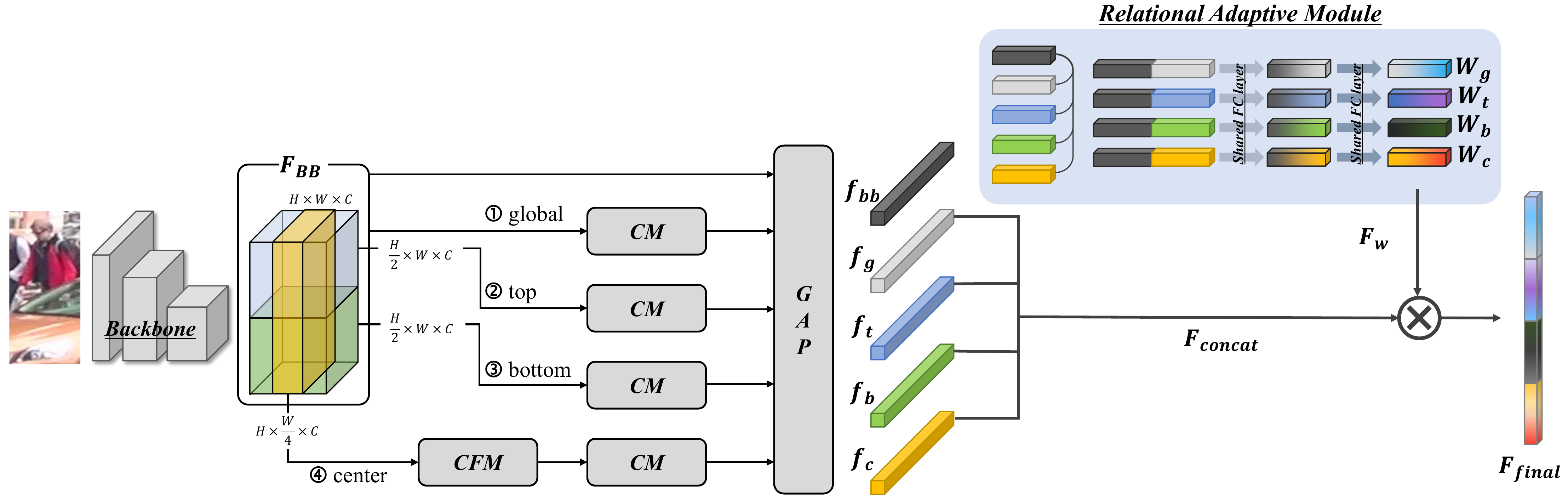}
		\caption{ Overall framework of the proposed approach. OCNet consists of Converter Module (CM), Center-Focus module (CFM), and Relation Adaptive Module (RAM). The backbone feature $\textbf{F}_{BB}$ go through CM to extract global features $\textbf{f}_g$, top features $\textbf{f}_t$, bottom features $\textbf{f}_b$, and center features $\textbf{f}_c$. After passing through RAM, it becomes a relational-weight $\textbf{F}_w$ with the same size as $\textbf{F}_{concat}$. Multiplying these two results in a final representation feature that is well-aligned and robust to occlusion. }
		\label{framework}
	\end{figure*}

Occlusion is classified into object occlusion and pedestrian-Interference (PI)~\cite{zhao2020not} based on the type of object that covers the target. In object occlusion cases, the global feature extracted by the network may be mixed with unnecessary information. In order to obtain only the target features, most methods~\cite{he2019foreground, miao2019pose} use attention maps or pose information generated by other additional networks such as pose estimation and human parsing networks. However, these methods are highly dependent on the performance of the external networks and are hard to train well with an error-prone pseudo label. Without such external help, it is impossible to flexibly deal with misalignment problems. The misalignment problem occurs when features are extracted in the order of important parts from the input image and simply concatenated. For example, in the Fig.~\ref{mis} (a), part features are extracted and concatenated in body-feet-upper body-head order and body-head-upper body-feet order for probe and gallery images; This leads to misalignment of the final representation features, lowering similarity and unmatching IDs.
Another issue is PI, which occurs when two or more people appear in the bounding box because of the limitations of detector performance in crowded scenes. As the network recognizes multiple pedestrians as the foreground, it is difficult to remove the non-target pedestrians. This phenomenon leads to mismatching between people and IDs. In existing methods~\cite{zhao2020not, he2020guided}, external networks are used or an attention map is obtained by performing additional operations upon query and gallery image to find the target.

To address the aforementioned issues, we propose Occlusion Correction Network (OCNet), a novel network that extracts global and various part features while performing feature correction in order to be robust to occlusion. Unlike the previous method~\cite{sun2018beyond} of simply concatenating the part features, which causes feature misalignment, our model predicts the weights through relational information between various features and corrects irrelevant information to create the final representation. Through relational-weight learning, the weights of the occluded part are adaptively determined and aligned with feature correction. Thus, our method extracts various features from the visible part of the given images even if occluded; it then characterizes the relationship between them to create a final representation feature that is robust to occlusion and contains a discriminative representation.

In this paper, our main contributions are as follows: (1) We design an \textbf{Occlusion Correction Network} that corrects the final representation through relational-weight learning to determine whether a part feature is crucial or not, so that a more diverse and representative feature is obtained which avoids occluded parts. (2) To provide a robust solution to pedestrian occlusion, we propose the essential concept of a \textbf{center feature} for the first time to provide a robust solution to pedestrian occlusion. (3) We suggest \textbf{Separation Loss} for focusing on different information between global features and part features. (4) Our network achieves \textbf{superior performance} not only in occluded dataset but also in holistic datasets, without relying on additionally trained networks or external annotations.
\begin{table*}[!t]
		\centering
		\caption{Comparison with state-of-the-art methods on occluded 
		Re-ID dataset~\cite{miao2019pose} and holistic Re-ID datasets~\cite{zheng2015scalable, zheng2017unlabeled, li2014deepreid}. Methods are divided into 4 groups: holistic-reid-targeted, attention-based, occlusion-targeted with external-cues-based and occlusion-targeted with no external-cues-based. The 2$^{\text{nd}}$ highest performance is underlined. "*" denotes different backbone.}
		\resizebox{1.75\columnwidth}{!}{
				\renewcommand{\arraystretch}{1}
				\begin{tabular}{c|cc||cc|cc|cc|cc}
					\hline
					\hline
					\multirow{2}*[-.3ex] {Method} &
					\multicolumn{2}{c||}{Occluded DukeMTMC} & 
					\multicolumn{2}{c|}{Market1501} & \multicolumn{2}{c|}{DukeMTMC-reID} &
					\multicolumn{2}{c|}{CuHK03-L} &
					\multicolumn{2}{c|}{CuHK03-D} \\
                    \cline{2-11} & Rank-1  & mAP & Rank-1 & mAP & Rank-1  & mAP & Rank-1 & mAP & Rank-1  & mAP\\
					\hline
					\hline
					 PCB \cite{sun2018beyond} (ECCV 2018) & 42.6 & 33.7 & 92.3 & 77.4 & 81.8 & 66.1 & - & - & 63.7 & 57.5 \\
					 BoT \cite{luo2019bag} (CVPRW 2019)& 47.7 & 41.3 & 94.1 & 85.7 & 86.4 & 76.4 & - & - & - & -\\
					 \hline
					 CASN \cite{zheng2019re} (CVPR 2019) & - & - & 94.4  & 82.8 & 87.7  & 73.7  & 73.7 & 68 & 71.5 & 64.4\\
					 CAMA \cite{yang2019towards} (CVPR 2019) & - & - & 94.7 & 84.5 & 85.8 & 72.9  & 70.1 & 66.5 & 66.6 & 64.2\\
					 \hline
					 PGFA \cite{miao2019pose} (ICCV 2019) & 51.4 & 37.3 & 91.2 & 76.8 & 82.6 & 65.5 & - & - & - & -\\
                     HoReID \cite{wang2020high} (CVPR 2020) & 55.1 & 43.8 & 94.2 & 84.9 & 86.9 & 75.6  & - & - & - & -\\
					 \hline
					 \textbf{OCNet (ours)} & 59.9 & 49.7 & 94.9 & 87.2 & 87.8 & 77.2 & \underline{77.9} & \underline{74.9} & \underline{76.7} & \underline{72.4}\\
					\hline
					\hline
					 ISP* \cite{zhu2020identity} (ECCV 2020) & 62.8 & 52.3 & \underline{95.3} & 88.6 & 88.7 & 78.9 & 76.5 & 74.1 & 75.2 & 71.4\\
					 MoS* \cite{jia2021matching} (AAAI 2021) & \textbf{66.6} & \textbf{55.1} & \bf{95.4} & \underline{89} & \bf{90.6} & 80.2 & - & - & - & -\\
					\hline
					\textbf{OCNet + ibn *(ours)} & \underline{64.3} & \underline{54.4} & 95 & \bf{89.3} & \underline{90.5} & \bf{80.2} & \bf{82} & \bf{78.6} & \bf{78.9} & \bf{76.2}\\
					\hline
					\hline
				\end{tabular}
			\label{t1}
		}
	\end{table*}
	
\section{PROPOSED METHOD}
\label{sec:format}
\begin{figure}[!t]
		\centering
		\subfloat[Converter Module
		]{\includegraphics[width=0.8\columnwidth]{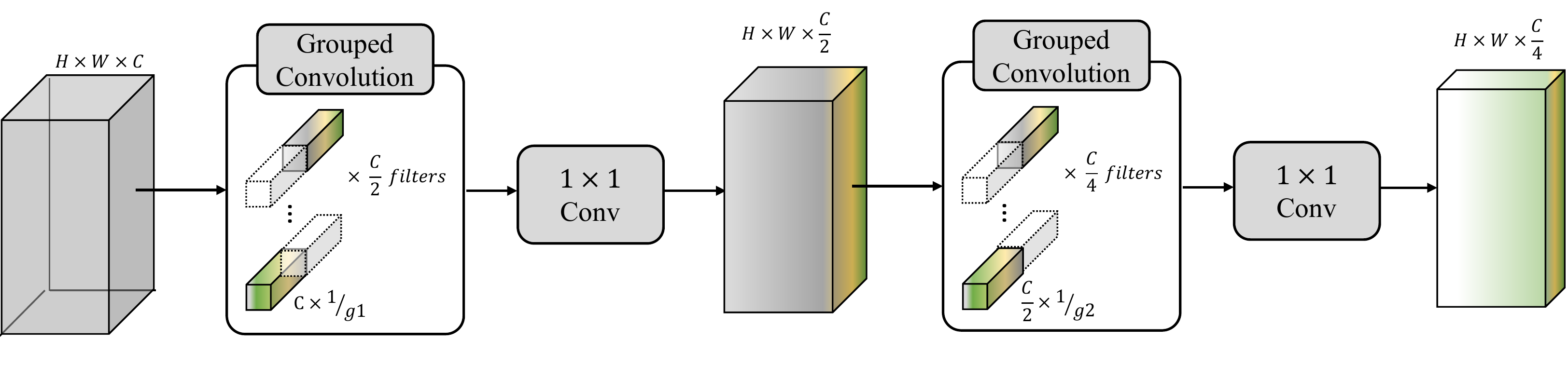}\label{CM}} \\
		\subfloat[Center-Focus Module ]{\includegraphics[width=0.8\columnwidth]{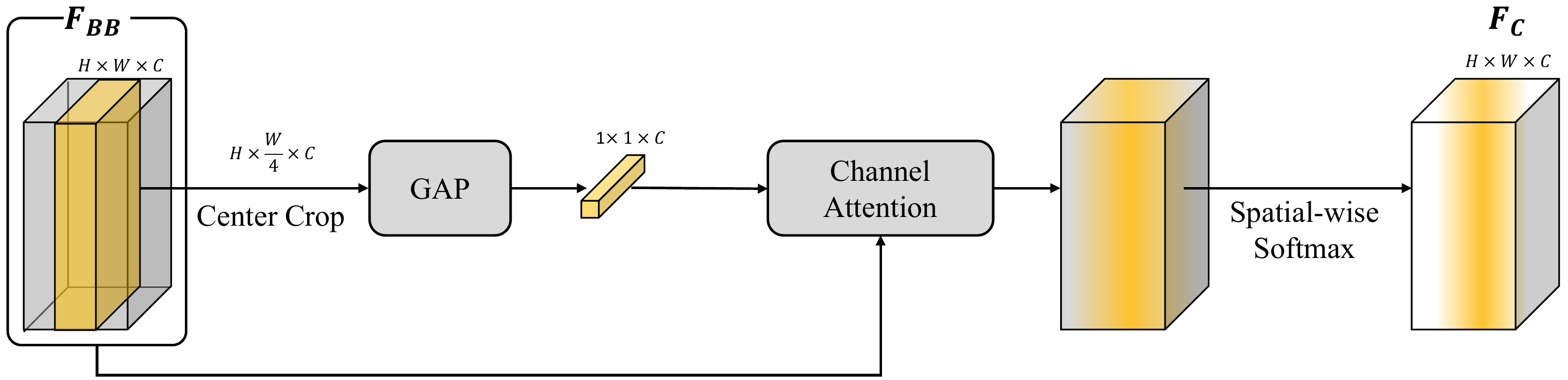} \label{CFM}}
		\\
		\caption{Overview of our (a) Converter Module (CM) (b) Center-Focus Module (CFM).} 
		\label{modules}
	\end{figure}
\subsection{Occlusion Correction Network}
\label{ssec:OCNet}
A practical solution for occluded Re-ID is to extract as many features as possible from the non-occluded regions of the given images. We extract four features ($\textbf{f}_g$, $\textbf{f}_t$, $\textbf{f}_b$, $\textbf{f}_c$) in various ways. Given that these features are generated from $\textbf{F}_{BB}$, which is activated in the foreground, they are not substantially affected by background cluttering. $\textbf{f}_g$ is obtained through a CM so that the network finds more important features in the existing $\textbf{F}_{BB}$. $\textbf{f}_t$ and $\textbf{f}_b$ are created by cropping the feature map extracted from the last layer of the backbone into two parts, top and bottom. OCNet learns through SL so that $\textbf{f}_t$ and $\textbf{f}_b$ can learn meaningful part-level features. $\textbf{f}_c$ generated by CFM is robust in PI because it focuses on the target. As shown in Fig.~\ref{framework}, we concatenate all these features ($\textbf{f}_g$, $\textbf{f}_t$, $\textbf{f}_b$, $\textbf{f}_c$) to form $\textbf{F}_{concat}$ and multiply the weight vector $\textbf{F}_w$ predicted by RAM. Therefore, we obtain the final representation feature $\textbf{F}_{final}$ in which the feature misalignment is solved.

\textbf{Converter Module.}
 For the part features of the pedestrian to have differently activated for each channel, a structure that divides and calculates a group of channels is needed. We adopt grouped convolution, which has the advantage of learning a channel with high correlation for each group. As shown in Fig.~\ref{modules} (a). CM not only reduces channels based on the high response of channels but also provides parameters that each feature learns with the corresponding objective function.

\textbf{Center-Focus Module.}
 In the PI problem, it is unclear which ID should be assigned as a label when the target is largely obscured by other people. In this case, using the detector mechanism, i.e., the natural assumption that the target is located in the middle of the bounding box, the feature focusing on the corresponding center part helps to find the target. CFM extracts features using the attention technique that weights the feature located in the middle of the bounding box. Unlike SENet~\cite{hu2018squeeze}, CFM crops the part corresponding to the center in the spatial domain of $\textbf{F}_{BB}$ and the softmax is taken to form a spatial-level probability map as shown in Fig.~\ref{modules} (b). Therefore, we extract a feature focusing on the center by multiplying and adding with $\textbf{F}_{BB}$. 

\subsection{Relational Adaptive Module}
\label{ssec:RAM}
 Most Re-ID methods~\cite{sun2018beyond, li2021diverse} concatenate various part features to make one final representation feature and use it for learning. However, it is not known what information each feature has, so a feature misalignment problem can occur when calculating the final feature distance. To solve feature misalignment, a correction process must be performed to estimate the contribution of each feature and reflect it to generate the final representation feature. 
We propose a RAM, which is different from relation network~\cite{santoro2017simple, park2020relation}, for estimating weights with relational information between features. 
(It is covered in \ref{Discussion}.) RAM consists of two fully connected shared layers that define relationships between vectors and handle occluded or insignificant features by weights. 

The $\textbf{F}_w$ generated in RAM is multiplied by $\textbf{F}_{concat}$ to finally correct the feature.

\subsection{Separation Loss Function}
\label{ssec:SL}
It is crucial to extract part-level features that are as diverse as possible from the parts representing the identity. We introduce a Separation Loss (SL) that prevents excessive focus on a specific region in an image with a single ID. SL calculates cosine similarity between ($\textbf{f}_g$, $\textbf{f}_t$) and ($\textbf{f}_g$, $\textbf{f}_b$) pairs as described in Eq.~\ref{SL}.
\begin{gather}
    L_{SL} = (1 + \frac{\textbf{f}_g\cdot \textbf{f}_t}{\left \| \textbf{f}_g \right \|_{2}{\left \| \textbf{f}_t \right \|_{2}}}) + (1 + \frac{\textbf{f}_g\cdot \textbf{f}_b}{\left \| \textbf{f}_g \right \|_{2}{\left \| \textbf{f}_b \right \|_{2}}})
    \label{SL}
\end{gather}
$\textbf{f}_t$ and $\textbf{f}_b$ are already uniformly horizontal partitions of $\textbf{F}_{BB}$. That is, they already have different information because they are physically separated. Therefore, SL learns to have different information because the value increases as $\textbf{f}_g$, which has global information, shares information with $\textbf{f}_t$ and $\textbf{f}_b$. 

\subsection{Training and Testing Phases}
\label{ssec:TT}
In the training phase, OCNet employs three kinds of losses, denoted as $L_{ID}$, $L_{Tri}$, and $L_{SL}$. As describes in Eq.~\ref{ID}, $L_{ID}$ and $L_{Tri}$ is the sum of $L_{ce}$ of $\textbf{F}_{final}$ , $\textbf{F}_{BB}$, $\textbf{f}_g$, and $\textbf{f}_c$ where $L_{ce}$ denotes cross-entropy with label smooth, and $L_{t}$ denotes triplet loss with margin, respectively. The overall objective function is defined as Eq.~\ref{Overall}. 
\begin{multline}
    L_{ID/Tri} = \lambda_{1}L_{ce/t}(\textbf{F}_{final}) + \lambda_{2}L_{ce/t}(\textbf{F}_{BB}) + \\ \lambda_{3}L_{ce/t}(\textbf{f}_g) + \lambda_{4}L_{ce/t}(\textbf{f}_c) 
    \label{ID}
\end{multline}
\begin{gather}
    L_{total} = L_{ID} + L_{Tri} + \gamma L_{SL}
    \label{Overall}
\end{gather}

In the testing phase, we calculate $l_2$ distances of $\textbf{F}_{final}$ and $\textbf{F}_{BB}$. Then, as in Eq.~\ref{test}, we consider both the distances of $\textbf{F}_{final}$ containing relational information and the distances of $\textbf{F}_{BB}$ containing general information from backbone.
\begin{gather}
D = \left \| \textbf{F}_{final}^{q} - \textbf{F}_{final}^{g} \right \|_{2} + \alpha \left \| \textbf{F}_{BB}^{q} - \textbf{F}_{BB}^{g} \right \|_{2}
    \label{test}
\end{gather}

\section{EXPERIMENTAL RESULTS}
\label{sec:pagestyle}
\textbf{Implementation Details:}
We adopt ResNet-50 as a backbone, pretrained using ImageNet. We refer to the structure in which a fully connected layer follows the batch normalization layer in and apply effective training strategies to use it as a baseline model~\cite{luo2019bag}. $\lambda{1}$, $\lambda{2}$, $\lambda{3}$, $\lambda{4}$ is set to 0.8, 0.5, 0.25, 0.25 respectively, depending on the importance of the feature. $\gamma$ and $\alpha$ are set to 1 in our experiments. 

\subsection{Comparison with State-of-the-art Methods}
\label{ssec:SOTA}
To show that our proposed method is effective not only in occlusion scenarios but also in general cases, we conduct experiments on two scenarios which is summarized in Table~\ref{t1}.

\textbf{Results on Occluded Re-ID dataset.}
 For a fair comparison, we compared with the 4$^{\text{th}}$ group using the ResNet-50 with ibn layer, which improves the generalization capacity of the model without increasing the model complexity. 1$^{\text{st}}$ group suffers when crucial information is occluded and 3$^{\text{rd}}$ group has no clear solution for PI, which leads to ID mismatching. OCNet has a lower performance than MoS but has the 2$^{\text{nd}}$ performance without additional operation in the inference. 

\begin{table}[!t]
		\centering
		\caption{Analysis of Relation Network (RN), concatenated features, and OCNet modules (CFM, SL, RAM).}
		\resizebox{1.0\linewidth}{!}{
				\renewcommand{\arraystretch}{1}
				\begin{tabular}{cc|ccccccccc}
					\hline
					\hline
					\multicolumn{2}{c|}{Index}&1&2&3&4&5&6&7&8&9\\
					\hline
				    \hline	
    \multicolumn{2}{c|}{RN}&&&&&\checkmark&\checkmark&&&\\
	\multicolumn{2}{c|}{Concat}&&&\checkmark&\checkmark&\checkmark&\checkmark&\checkmark&\checkmark&\checkmark\\
				\hline	\multirow{3}[1]{*}{\begin{sideways}\textbf{OCNet}\end{sideways}}&\multicolumn{1}{|c|}{CFM}&&\checkmark&&&\checkmark&\checkmark&\checkmark&&\checkmark\\
					&\multicolumn{1}{|c|}{SL}&&&&\checkmark&&\checkmark&&\checkmark&\checkmark\\
					&\multicolumn{1}{|c|}{RAM}&&&&&&&\checkmark&\checkmark&\checkmark\\
					\hline
					\multicolumn{2}{c|}{Rank-1}&47.7&49.2&55.2&58.6&52.7&53&52.9&58.6&\textbf{59.9}\\
					\multicolumn{2}{c|}{mAP}&41.3&42&45.9&49.1&44.2&43.2&45.1&49.5&\textbf{49.7}\\
					\hline
					\hline
				\end{tabular}}
				\label{t2}
\end{table}
	
\textbf{Results on Holistic Re-ID datasets.}
 Even though our OCNet is occluded-targeted method, it outperforms the most recent state-of-the-art method on holistic datasets. In particular, in CUHK03-D, the bounding box is not tight, so multiple people often appear in the image. Our method exceeds state-of-the-art methods in the dataset corresponding to the problem we are trying to solve.
	
\subsection{Discussion}
\label{Discussion}
We conduct extensive ablation studies on Occluded-Duke MTMC. It is summarized in Table~\ref{t2}.

\textbf{OCNet.} When comparing Index-1 and Index-3, extracting various features enables us to discriminatively express identity. Index-9, where all of our proposed modules are applied, shows a significant difference in performance. 

\textbf{CFM \& SL.} Index-2 extracts features focused on the center by CFM. Index-8 removes CFM from OCNet. When comparing Index-1 and Index-2, extracting the feature through CFM helps to find the target. From index-7 and index-9, it indicates that SL enables the extraction of various features.

\textbf{RAM.}
Index-5 learns using vectors containing relational information by applying the existing Relation Network (RN)\cite{santoro2017simple}. A comparison of index-6 and index-9 shows the difference between our RAM and the RN. If the relation vectors are used in occlusion scenarios, unnecessary information is intertwined. By contrast, RAM predicts weights based on relational information and corrects features to generate discriminative features that are robust against occlusion.

\textbf{Influence of the Size of Center.}
From Table~\ref{t3}, the best performance is when $W$ is 2. This is because, contrary to the intention of CFM, as the size of the feature map to focus increases, information about non-target objects gets mixed in. 

\textbf{Influence of the Number of Parts.}
 The more parts there are, the more modules need to be added, and the size of the feature also increases. Considering this, we set it to two parts. 

\textbf{Analysis of Hyperparameters of SL.}
From Table~\ref{t3}, when $\gamma$ is 1, the best performance is achieved. SL must be smaller than the other losses to avoid interfering with the identification objective.

\begin{table}[!t]
		\centering
		\caption{Comparison with varing settings of hyperparameters on Occluded-DukeMTMC.}
		\resizebox{1\linewidth}{!}{
				\renewcommand{\arraystretch}{1}
				{\huge
				\begin{tabular}{c|cccc||cccc||cccc}
					\hline
					\hline
					Hyper- &
					\multicolumn{4}{c||}{Size of Center} & 
					\multicolumn{4}{c||}{Num of Parts} & \multicolumn{4}{c}{$\gamma$} \\
                    \cline{2-13}param. & 2 & 4 & 6 & 8 & 1 & 2 & 3 & 4 & 0 & 1 & 2 & 5\\
					\hline
					\hline
					 Rank-1 & \textbf{59.9} & 58.1 & 59.2 & 59 & 57.8 & \textbf{59.9} & 59.4 & 59.5 & 52.9 & \textbf{59.9} & 57.7 & 55.6\\
					 mAP & \textbf{49.7} & 49.3 & 49.4 & 49.6 & 48.2 & \textbf{49.7} & 49.2 & 49.9 & 45.1 & \textbf{49.7} & 48.5 & 47\\
					 \hline
					 \hline
				\end{tabular}}}
				\label{t3}
\end{table}

\begin{figure}[t]
		\centering
		\includegraphics[width=1.0\columnwidth]{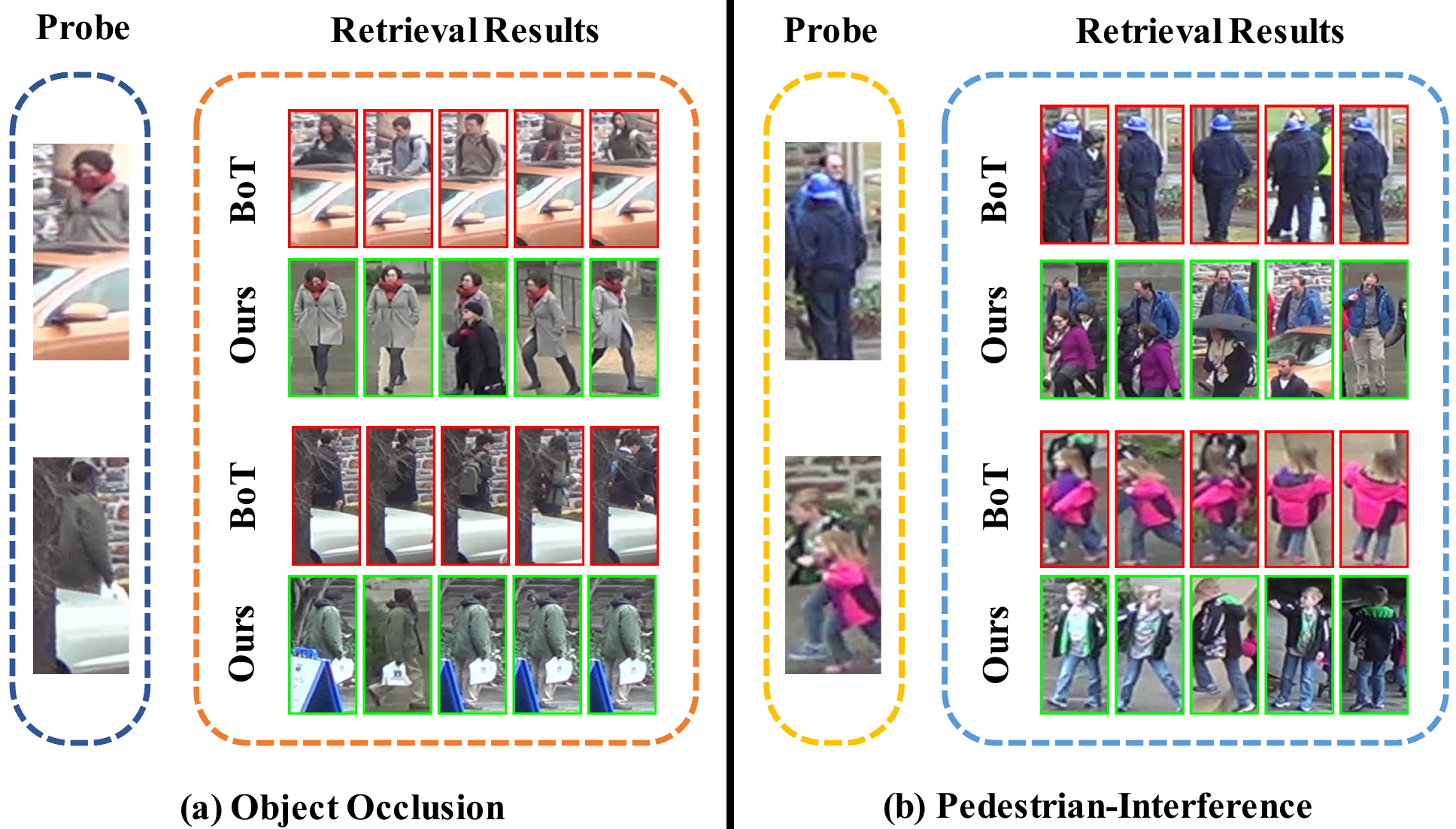}
		\caption{Ranking results of BoT and OCNet. The red and green bounding indicate the error and correct result, respectively.}
		\label{visual}
	\end{figure}

\subsection{Visualization}
\label{ssec:Vis}
As shown in Fig.~\ref{visual} (a), BoT~\cite{luo2019bag} has a tendency to recognize the obstacle as a part of the feature, so images where the obstacle and the target coexist, are retrieved. However, OCNet which corrects the occluded part feature successfully retrieval by concentrating only on the features corresponding to the target. The Fig.~\ref{visual} (b) shows pedestrian-interference samples, where the non-target occludes the target significantly. BoT is biased towards non-targets that are more visible than the target. On the contrary, OCNet successfully retrievals the target.

\section{CONCLUSION}
\label{sec:typestyle}
In this paper, we propose an Occlusion Correction Network (OCNet) that corrects features through relational-weight learning and obtains diverse and representative features without extra cues. Our method generates features that are robust against object occlusion as well as pedestrian interference. Our network also has an exceptional generalized ability that shows good performance even in holistic Re-ID.

\vfill\pagebreak

\bibliographystyle{IEEEbib}
\bibliography{strings}

\end{document}